\documentclass[a4paper,12pt]{esslli}

\usepackage[dvips]{graphicx} 
\usepackage{gb4e} 
\usepackage{cgloss4e} 
\usepackage{a4wide} 
\usepackage{chicago} 

\begin{document}

\chapter{DepAnn - An Annotation Tool for Dependency Treebanks}   

\thispagestyle{firstpage}

\chapterauthor{Tuomo Kakkonen} 
\chapteraffiliation{Department of Computer Science, University of Joensuu, Finland}
\chapteremail{tuomo.kakkonen@cs.joensuu.fi}

\begin{chapterabstract}
DepAnn is an interactive annotation tool for dependency treebanks, providing both graphical and text-based annotation interfaces. The tool is aimed for semi-automatic creation of treebanks. It aids the manual inspection and correction of automatically created parses, making the annotation process faster and less error-prone. A novel feature of the tool is that it enables the user to view outputs from several parsers as the basis for creating the final tree to be saved to the treebank. DepAnn uses TIGER-XML, an XML-based general encoding format for both, representing the parser outputs and saving the annotated treebank. The tool includes an automatic consistency checker for sentence structures. In addition, the tool enables users to build structures manually, add comments on the annotations, modify the tagsets, and mark sentences for further revision.
\end{chapterabstract}


\section{Introduction}

Treebanks, collections of syntactically annotated sentences, are needed for developing and evaluating \emph{natural language processing} (NLP) applications, as well as for research in empirical linguistics. The earliest treebanks, constructed in 1970's, were annotated manually \cite{abe:03}. As treebank construction is labor-intensive, methods are needed for automating part of the work. The reason that treebanks are not constructed fully automatically is obviously the fact there are no parsers of free text capable of producing error-free parses. In semi-automatic treebank building, the work of an annotator is transformed from a tree builder to a checker and corrector of automatically created structures. Constructing a treebank semi-automatically calls for a range of tools, such as a \emph{part-of-speech} (POS) tagger, a syntactic parser and an annotation tool.

In recent years, there has been a wide interest towards dependency-based annotation of treebanks. Dependency grammar formalisms stem from the work of Tesni\'{e}re \shortcite{tes:59}. Most often the motivation for basing the treebank format on dependency is the fact that the language for which the treebank is developed for has a relatively free word order. In such languages, due to their rich morphology, there is more freedom in word order for expressing syntactic functions. In dependency-based grammars, only the lexical nodes are recognized, and the phrasal ones are omitted. The lexical nodes are linked with directed binary relations. The dependency structure of a sentence thus consists of a number of nodes which is equal to the number of words in the sentence, a root node and the relations (dependency links) between the nodes. 

Although more collaboration has emerged between treebank projects in recent years, the main problem with current treebanks in regards to their use and distribution is the fact that instead of reusing existing annotation and encoding schemes, new ones have been developed. Furthermore, the schemes that have been developed are often designed from theory and even application-specific viewpoints, and consequently, undermine the possibility for reuse. In addition to the difficulties for reuse, creating a treebank-specific representation format requires developing a new set of tools for creating, maintaining and searching the treebank.

The main motivation for designing and implementing \emph{DepAnn} (Dependency Annotator), an annotation tool for dependency treebanks, stems from the need to construct a treebank for Finnish. As Finnish is a language with relatively free word order, dependency-based annotation format is a straight-forward choice as the basis for the annotation. Although DepAnn is customized to be used for creating the Finnish treebank, the choices made in the architecture and design of the system allow it to be modified to the needs of other treebank projects. Most importantly, DepAnn uses a XML-based abstract annotation format, TIGER-XML \cite{men:00} as both input and output formats.

This paper represents the main design principles and functionality of DepAnn. In addition, we describe how the system interacts with the other treebanking tools (POS taggers, morphological analyzers, and parsers). Section 2 shortly describes the principles of treebank construction. Section 3 represents the requirements defined for DepAnn based on an analysis of existing annotation tools, and describes the tool. Finally, in Section 4 we give concluding remarks and underline some future possibilities.

\section{Background}
Speed, consistency, and accuracy are the three key issues in treebank annotation. The most commonly used method for constructing a treebank is a combination of automatic and manual processing. Constructing a treebank, even with a semi-automatic method, is a labor-intensive effort. Efficient tools play a key role in lowering the costs of treebank development and enable larger, higher quality treebanks to be created. Both goals are crucial. The estimated costs of the Prague Dependency Treebank, the largest of the existing dependency treebanks, are USD 600,000 \cite{boh:03}. A treebank has to be large enough to have any practical use, for example for grammar induction. The size of the existing dependency treebanks is quite limited, ranging from few hundreds to 90,000 sentences. Self-evidently, a treebank has to be also consistent and have a low error frequency to be useful. 

A morphological analyzer and a parser should be applied in order to lower the burden of the annotators. The typical procedure is to use a parser that leaves at least part of ambiguities unresolved and dependencies unspecified, and let human annotators to do the inspection and correction of the parses. Thus, an annotator is correcting the POS and morphosyntactic tags, resolving the remaining ambiguities and adding and correcting any missing or erroneous dependencies. A crucial component in this type of semi-automatic treebank creation is the annotation tool. A well-designed and well-implemented tool can aid the work of annotators considerably. With an annotation tool, the user can browse, check, and correct the parser's output as well as create structures from scratch. In some of the existing tools the annotations are automatically checked against inconsistencies before saving them to the treebank. In addition, the user is able to add comments to the structures or mark them as doubtful.

Dependency treebanks have been built for several languages, \emph{e.g.} Czech \shortcite{boh:03}, English \shortcite{ram:02}, Danish \cite{bick:03,kro:03}, Italian \cite{les:02}, and Dutch \cite{van:02}. The \textit{TIGER Treebank} of German is an example of a treebank with both phrase structure and dependency annotations~\cite{bra:02}. The current direction in the thinking in the dependency vs. constituency discussion in general is on integration and cooperation \cite{sch:98}. While dependency grammars are superior in handling free word order, on one hand some elements of constituency grammars are better for handling certain phenomena (\emph{e.g.} coordination), and on the other hand, constituency-based grammars also need dependency relations, at least for verb valency. Furthermore, dependency structures can be automatically converted into phrase structures \cite{xia:01} and vice versa \cite{daum:04}, although not always with 100\% accuracy.

We started designing a treebank for Finnish by analyzing the methods and tools used by other dependency treebank projects. The producers of the dependency treebanks have in most cases aimed at creating a multipurpose resource for research on NLP systems and theoretical linguistics. Some, \emph{e.g.} the \emph{Alpino Treebank} of Dutch \shortcite{van:02}, are built for a specific purpose. Most of the dependency treebanks consist of newspaper text and are annotated on POS, morphological and syntactic levels. An interested reader is referred to \cite{kak:05} for further details on the analysis of dependency treebanks.

After a throughout study of existing annotation methods and tools (such as \emph{GRAPH} \cite{boh:03}, \emph{Abar-Hitz} \cite{dia:04}, \emph{Annotate} \cite{pla:00}, \emph{DTAG} \cite{kro:03}), \emph{CDG SENtence annotaTOR} (SENATOR) \cite{whi:00}, it was found that none of the available annotation tools satisfied all our requirements. Some tools were not suitable for dependency annotation, some were not compatible with any common XML-based annotation formats, the user-interface was not considered suitable or the tool didn't have all the functions we required. In addition, to our knowledge there aren't any annotation tools available capable of showing or merging outputs from several parsers for aiding the annotator's choices. Thus, the decision was made to design and implement an annotation tool with all the desired characteristics.

\section{The annotation tool}

\subsection{Design principles}

The analysis of existing annotation tools was crucial in defining the requirements for the system to be developed. The following key features were recognized:

\begin{itemize}

\item\emph{Support for an existing XML encoding scheme}\newline
Building a treebank is such a labor-intensive effort that promoting co-operation between treebank projects and reuse of formats and tools is an important and widely accepted goal in treebanking community (\emph{e.g.} \cite{ide:03}). Using an existing encoding format will make the system reusable. In addition, existing tools supporting the same scheme can be used for browsing, manipulating and searching the annotated treebanks.

\item\emph{Both textual and graphical display and manipulation of parse trees}\newline
For any annotation tool the capability to visualize the sentence structures is a necessity. In addition, the graphical view should preferably be interactive, so that the user can manipulate the structures. On the other hand, for some annotation tasks or for some user's needs textual view of the structure may be more suitable.

\item\emph{An interface to morphological analyzers and parsers for constructing the initial trees}\newline
In order to generate the initial trees for human inspection and modification, the annotation tool must have an interface to a morphological parser, a POS tagger and a syntactic parser. The tool should be able to use simultaneously outputs from several tools to guide the annotator's decisions.

\item\emph{An inconsistency checker for both structures and encoding}\newline
The annotated sentences to be saved to the treebank should be checked against tagging inconsistencies. In addition to XML-based validation of encoding, the inconsistency checker should inform the annotator about several other types of mistakes, such as mismatching combinations of POS and morphological tags, missing main verb, and fragmented, incomplete parses.

\item\emph{Menu-based tagging}\newline
In order to make the annotation process faster, setting the tags should be done by means of selecting the most suitable tag from a pre-defined set of tags, instead of requiring the annotator to type the tag label. In addition to being efficient, menu-based tagging lowers the number of errors as there will be no errors cost by typos in the labels. On the other hand, keyboard shortcuts for selecting appropriate tags should be provided for more advanced users.

\item\emph{A commenting tool}\newline
For easing the later revisions, possibly performed by other annotators, the user should be able to add comments on the annotated structures. In addition, user should be able to mark a sentence as ready or unfinished to make it easier to locate sentences needing further revision.

\end{itemize}

The foremost design principle, apart from making the annotation process faster and less error-prone, was that the tool must be reusable and modifiable. The system was designed in way that the modules for processing the treebank output and input are kept separate from the structure viewing and manipulation modules, thus making the tool more easy to modify. The support for an existing encoding scheme is a crucial reusability feature of any treebanking software. The selection of the format was first narrowed down by the decision that the format should be XML-based, as XML offers a set of validation capabilities, in order to automatically check for encoding inconsistencies.

The aim of an abstract annotation model is to provide a general framework for linguistic annotation. Existing abstract annotation formats share the common goal of offering an intermediate level between the actual data (encoding scheme) and the conceptual level of annotation (annotation scheme). An advantage of such an approach is to enable a common set of tools to be used for creating and manipulating treebanks in several formats. From the set of possible option, including \emph{e.g.} XCES \cite{ide:03}, TIGER-XML \cite{men:00} was selected to be used in DepAnn. TIGER-XML is an exchange format for corpora and treebanks, providing an XML-based representation format which is general enough for representing diverse types of corpus and treebank annotations \cite{men:00}. The format is based on encoding of \emph{directed acyclic graphs} (DAGs). Each DAG represents a sentence as terminal (\emph{i.e.} words) and nonterminal (dependencies) nodes. The syntactic categories, POS, lemma and other information is represented as attributes in the nodes. The edges encode labeled links between terminals and nonterminals.

TIGER-XML has several desirable characteristics: First, it is flexible and extensible enough to accommodate different treebank annotation types, both dependency and consistency based. Second, it has been shown to be suitable for dependency annotation in several treebank projects (\emph{e.g.} TIGER Treebank \shortcite{bra:02}, Arboretum \cite{bick:03}). Third, there are explicit specifications available how to encode dependency structures in the scheme \cite{kro:04}. And finally, there exists a set of well-implemented tools supporting the format, such as \emph{TIGERSearch} viewing/query tool and \emph{TIGERRegistry} indexing tool \cite{kon:03}, capable of transforming some well-known corpus and treebank formats, such as the \emph{SUSANNE} \cite{sam:95} and \emph{Penn Treebank} \cite{mar:93} into TIGER-XML. 

As TIGER-XML is a general model of treebank encoding, it would be possible to show and manipulate constituency structures with DepAnn. However, the decision was made that the tool was not going to be designed for both constituent and dependency structures in a suspicion that too general design would hamper the efficiency of dependency annotation. Thus, the visualization functions and the user interface are tuned for manipulating dependency structures.

\subsection{Main functionality}

In DepAnn tool, the structure to be annotated is represented to the user in textual and graphical formats in order to offer the best option for each user's needs. The textual and graphical views are fully integrated, thus the changes applied in the graphical view immediately affect the textual one and vice versa. The user interface is customizable to suit the task and the annotator's preferences. The user can add comments on annotations, reminding on problematic parts on the sentence structures. Completed trees can be marked as ready, indicating that no further inspection and modifications are needed.

Outputs of several parsers and POS taggers can be applied in parallel to offer the annotator a possibility to compare the outputs in order to guide the annotation decisions. To be able to use the output of an parser in DepAnn, a converter must be implemented to transform the output from the parser or tagger-specific format to the format used by DepAnn. TIGER-XML \cite{men:00} is used as the input format for the structures obtained from the automatic tools, as well as the output format for the annotated treebank. For internal data representation the TIGER-XML structures are transformed into Java objects. Figure \ref{figure1} illustrates the input and output processes of DepAnn. 

\begin{figure}
\centering
\caption{
The inputs and outputs of the tool.
\label{figure1} 
}
\end{figure}

The annotation process using DepAnn starts with processing the treebank texts with one or more parsers and taggers. Next, a converter is applied to the outputs in order to transform the tool-specific format into TIGER-XML. After the conversion, the annotator can view the parsed structures and build the annotated structure to be added to the treebank. The user can select the parser output to be used for creating the initial trees. Figure \ref{figure2} illustrates the main frame of DepAnn's user interface.

\begin{figure}[ht]
\centering
\caption{ 
\label{figure2} 
The main frame of DepAnn tool.
}
\end{figure}

The main groups of functions are indicated in Figure \ref{figure2} by boxes A...E. The text field in the area bordered with box A shows the sentence being annotated in raw text format. Area B is a toolbar with controls for treebank browsing (buttons for showing the next and the previous sentence and a slidebar for browsing), checking and saving the sentence, and modifying the tag sets. In C, the user can graphically manipulate the structure by changing the values on nodes representing the words and dependency links and by removing, adding and rerouting the links between the nodes. Area D consists of the revision functions. User can mark the sentence as ready, indicating that further revision is not needed. In addition, user can use the comment field to write notes concerning the sentence structure. Box E frames the tables for text-based structure manipulation and viewing.

The parser and tagger outputs for aiding the annotation decisions are shown in a separate resizable, customizable dialog. For example, in a computer system with multiple monitors, the dialog can be placed in to a separate desktop. In the current version, the user can select which parser's output is used as the initial tree for correction and modification. We are working on an extension to the system, in which the initial trees would be created by semi-automatically combining the parsers' and taggers' outputs by the aid of the annotator.

When the user decides to stop editing a sentence, an automatic consistency checking is performed to validate the sentence structure, the annotation, and encoding. First, a series of checks are run to verify that the sentence has a main verb, a root, all the words have word form and lemma information and morphosyntactic tags, and that the sentence is not fragmented \emph{etc}. Second, if the first series of checks was passed, the sentence is transformed into TIGER-XML and validated against the XML schema to find any errors in encoding. The problems found are indicated to the user. The user can select which checks are run by modifying the system set-up.

\subsection{Implementation details}

The annotation tool is implemented in Java. As Java is platform-independent, the system can be used in any environment for which Java is available. The system consists of three main components: the interface to parsers and taggers, the annotation tool itself, and the output module. Two freely available open source packages, \emph{OpenJGraph} \cite{sal:06} and \emph{TIGER API} \shortcite{dem:06}, were used for developing the system, although both had to be modified considerably to be suitable to be used as a part of DepAnn. TIGER API, a Java API for TIGER-XML, is used for input and output processing. The graphical annotation manipulation functionality was build on top of OpenJGraph. The annotation tool uses \emph{Java Database Connectivity} (JDBC) for storing the outputs from the parsing and tagging tools, as well as for the user comments and information on ready sentences. Thus, the MySQL database currently in use can be replaced by any other JDBC-compatible database.

\section{Conclusion}
The semi-automatic annotation tool for dependency structures discussed in the paper provides graphical and text-based annotation functions, possibility to use outputs from several parsers to aid the annotation decisions, tools for commenting the annotated structures, automatic consistency checking, and support for TIGER-XML format. In its first application, DepAnn will be used for creating a treebank for Finnish, aimed for evaluation of syntactic parsers. Outputs from two parsers/morphological analyzers, \emph{Functional Dependency Grammar} parser (FI-FDG) \cite{tap:97} and \emph{Constraint Grammar} parser (FINCG) \cite{kar:90} is transformed to TIGER-XML and represented to the annotator as the basis for creating the correct structure. The tool is implemented in a way that it is adjustable for other treebank projects' needs. As the annotation format is based on TIGER-XML, the tool is not restricted to a particular set of POS, morphological or dependency tags. The modules for processing the treebank output and input are separate from the graphical and textual annotation modules, thus the tool could be modified to use any other annotation format. DepAnn will be made publicly available as an open source distribution. 

As mentioned above, the issues related to reuse of tools and formats is one of the major issues in treebanking. Thus, few words on development costs of the annotation tool is in order. The work was conducted by a researcher with a degree in Software Engineering and few years of practical experience in programming and software designing. No exact data was recorded, but the amount of work to design and implement the system to its current state is around a half of a man-year. The work was considerably eased by using open source APIs for treebank manipulation and graph visualization. These observations underline the importance of reusing existing annotation schemes and software components for treebank development.

As discussed earlier, an improvement to the system that we are currently working on is the semi-automatic creation of initial trees. The algorithm would automatically combine as many words and dependency links of the taggers' and the parsers' outputs as possible, and ask the annotator the make decisions on the rest. Such method would improve the quality of the initial trees, thus lowering the number of modifications needed to come up with the correct structure. Other future enhancements to the system could include even more strict and detailed checking algorithms for the annotated structures and an improved interface between DepAnn and the parsers which would allow the annotator to interact with the parsers in a case of problematic sentences. The approach has been successfully applied by some annotation tools, such as Annotate \cite{pla:00} and the lexical analysis and constituency marking tools of the Alpino Treebank \cite{van:02}. Often several annotators are working on the same sentences in order to ensure the consistency of the treebank.  In such cases, it would be helpful if the tool would allow to manage multiple annotations and to perform inter-annotator agreement checks. Furthermore, the memory management of the tool could be improved in order to make it more efficient when working with large treebanks with tens of thousands of sentences.

\section*{Acknowledgements} 
The author would like to thank the two anonymous reviewers for helpful comments on this paper and for some interesting suggestions for future work. The research reported in this paper has been supported by the European Union under a Marie Curie Host Fellowship for Early Stage Researcher Training at MULTILINGUA, University of Bergen, Norway, MirrorWolf project funded by the National Technology Agency of Finland (TEKES), and Automated Assessment Technologies for Free Text and Programming Assignments project funded by the Academy of Finland. The work was partly conducted while the author was working at the Human Language Technology Group at the Council for Scientific and Industrial Research (CSIR), Pretoria, South Africa and at the Faculty of Philosophy, University of Split, Croatia. 

\bibliographystyle{chicago} 
\bibliography{kakkonen} 

\end{document}